\documentclass[11pt,a4paper]{article}
\usepackage[margin=2.2cm]{geometry}
\usepackage[T1]{fontenc}
\usepackage{lmodern}
\usepackage{graphicx}
\usepackage{amsmath,amssymb,amsfonts}
\usepackage{booktabs}
\usepackage{array}
\usepackage{enumitem}
\usepackage{xcolor}
\usepackage{csquotes}
\usepackage[numbers,sort&compress]{natbib}
\usepackage{hyperref}
\usepackage{tikz}
\usetikzlibrary{arrows.meta,positioning,fit,calc}

\definecolor{darkgray}{HTML}{333333}
\definecolor{midgray}{HTML}{666666}
\definecolor{lightgray}{HTML}{F2F2F2}
\hypersetup{
  colorlinks=true,
  linkcolor=darkgray,
  urlcolor=blue,
  citecolor=darkgray,
  pdfauthor={Jacques Raynal; Pierre Slangen; Elsa Raynal; Jacques Margerit},
  pdftitle={Detecting Explanatory Insufficiency in Learned Representations: A Framework for Representational Vigilance}
}
\setlist{itemsep=0.25em,topsep=0.35em}

\newcommand{\conceptbox}[1]{%
\begin{center}
\fcolorbox{darkgray}{lightgray}{\parbox{0.91\textwidth}{\vspace{0.35em}#1\vspace{0.35em}}}
\end{center}}

\tikzset{
  box/.style={draw=darkgray, rounded corners=2pt, thick, align=center, fill=lightgray, inner sep=6pt, font=\small},
  arrow/.style={-{Latex[length=2.2mm]}, thick, draw=midgray},
  dashedarrow/.style={-{Latex[length=2.2mm]}, thick, dashed, draw=midgray},
  label/.style={font=\scriptsize\itshape, text=midgray, align=center}
}

\begin{document}

\begin{center}
{\LARGE\bfseries Detecting Explanatory Insufficiency in Learned Representations:\\[0.2em]
A Framework for Representational Vigilance}\par
\vspace{0.8em}
{\large Jacques Raynal$^{1,*}$, Pierre Slangen$^{2,3}$, Elsa Raynal$^{4}$, Jacques Margerit$^{5}$}\par
\vspace{0.55em}
\small
$^{1}$Laboratory of Bioengineering and Nanosciences (LBN), Universit\'e de Montpellier, Montpellier, France\\
$^{2}$EuroMov Digital Health in Motion, Universit\'e de Montpellier, Montpellier, France\\
$^{3}$IMT Mines Al\`es, Al\`es, France\\
$^{4}$Sensorimotor Practice, Montpellier, France\\
$^{5}$Universit\'e de Montpellier, Montpellier, France\\[0.35em]
$^{*}$Corresponding author: \href{mailto:jacques.raynal@umontpellier.fr}{jacques.raynal@umontpellier.fr}\\
ORCID: \href{https://orcid.org/0000-0002-0604-2733}{0000-0002-0604-2733}\\[0.35em]
\textit{Substantial revision prepared 21 September 2026}
\end{center}

\begin{abstract}
Learned representations are usually evaluated through predictive performance, calibration, robustness, uncertainty estimation, or behavior under distribution shift. These criteria are indispensable, but they need not determine whether the representation itself remains sufficient for organizing the observations that matter to an explanatory task. This paper develops VER (from the French \emph{Vigilance Explicative et Représentationnelle}; hereafter explanatory and representational vigilance) around a narrower scientific question: can a system detect when its current representation has become explanatorily insufficient without confusing insufficiency with ordinary error, missing data, uncertainty, or distribution shift? The current operational form of VER is organized as TRACE $\rightarrow$ EVIDENCE $\rightarrow$ INSUFFICIENCY $\rightarrow$ REGIME HYPOTHESES $\rightarrow$ TRANSITION ASSESSMENT $\rightarrow$ VIGILANCE $\rightarrow$ PROBE. The framework treats structured residuals as signals rather than verdicts, distinguishes data insufficiency from representation insufficiency, preserves multiple hypotheses under non-identifiability, allows a legitimate NON-DETERMINED outcome, and requires a discriminating Probe before stronger conclusions are drawn. It also separates variation, drift, adaptation, and representational transition, and introduces a Present Enrichment test before attributing explanatory value to history. VER is positioned relative to model criticism, uncertainty estimation, out-of-distribution detection, mechanistic interpretability, computational reliabilism, and optimal experimental design. Two paired thought experiments and a falsifiable benchmark are proposed. The framework is not claimed to be empirically validated here; its contribution is an explicit architecture designed for auditability, intended to determine when representational adequacy deserves explicit investigation and what observation would most directly discriminate among competing explanations.
\end{abstract}


\conceptbox{\textbf{Revision note.} Earlier public versions of this manuscript described VER through five diagnostic operations, a regime-aware assessment gate toward representational transition, and recursive vigilance across successive representations. This revision supersedes that operational architecture and aligns the article with the current VER specification: TRACE $\rightarrow$ EVIDENCE $\rightarrow$ INSUFFICIENCY $\rightarrow$ REGIME HYPOTHESES $\rightarrow$ TRANSITION ASSESSMENT $\rightarrow$ VIGILANCE $\rightarrow$ PROBE. Regime-aware representational transition and recursive cross-level history are no longer treated as components of VER itself; they are downstream or neighboring questions for explicit reasoning over alternative representations and representational change. The purpose of the revision is to sharpen scope, falsifiability, and the distinction between vigilance and revision.}

\section{Introduction}

Learned representations support classification, prediction, generation, planning, transfer, and adaptation across modern machine learning systems \citep{Bengio2013,Goodfellow2016}. Their quality is usually assessed through downstream performance, calibration, robustness, uncertainty estimation, or behavior under distribution shift \citep{Guo2017,Ovadia2019,Hullermeier2021,HendrycksGimpel2017}. These criteria are essential. Yet a system can remain operationally useful while a structured subset of observations is repeatedly difficult to integrate, distinguish, or explain within the representation currently organizing the task.

This possibility is not equivalent to the familiar claim that all models are imperfect. Statistical model criticism has long emphasized residual checking, mismatch, and revision \citep{Box1976,GelmanMengStern1996,GelmanShalizi2013}. Nor is it equivalent to uncertainty estimation, out-of-distribution detection, or interpretability. A model can be uncertain while its representation is adequate for the task; it can be confident while a systematic explanatory ambiguity persists; and it can be interpretable without being sufficient for the observations that matter. Likewise, a reliable computational process can justify routine reliance while still producing a locally discordant result that deserves additional examination \citep{DuranFormanek2018,SchmidtBlatt2026,Ferrario2026}.

The scientific problem addressed here is therefore narrower than general model evaluation:

\conceptbox{\textbf{Canonical question.} \emph{Is the representation currently being used sufficient to explain the observations available for the declared task?}}

For learned systems, this question becomes operationally difficult because output error, data insufficiency, distribution shift, latent variables, non-identifiability, and representational inadequacy can produce superficially similar symptoms. A useful framework must therefore do more than detect anomalies. It must separate what is observed from what is inferred, keep alternative explanations alive when the data do not identify a unique account, and select a new observation or intervention that can discriminate among them.

This paper develops \textbf{VER} (from the French \emph{Vigilance Explicative et Représentationnelle}; hereafter \emph{explanatory and representational vigilance}). VER does not propose a new learning algorithm, loss function, or architecture. It is a diagnostic and experimental-orchestration layer. Its current operational form consists of seven stages:

\[
\begin{gathered}
\text{TRACE} \rightarrow \text{EVIDENCE} \rightarrow \text{INSUFFICIENCY} \rightarrow \text{REGIME HYPOTHESES}\\
\rightarrow \text{TRANSITION ASSESSMENT} \rightarrow \text{VIGILANCE} \rightarrow \text{PROBE}
\end{gathered}
\]

The contribution is not that residual analysis, non-identifiability, model checking, or informative experimental design are individually new. They are not. The proposal is that these established concerns can be organized around \emph{representation sufficiency} as the explicit object of monitoring, with mandatory safeguards against premature diagnosis and a final requirement for a discriminating Probe. This makes the framework falsifiable: if VER adds no useful distinctions, produces excessive false alerts, or proposes Probes no more discriminating than standard baselines, the stronger methodological claim should be rejected.

The paper proceeds as follows. Section~\ref{sec:problem} defines explanatory insufficiency and distinguishes it from neighboring failure modes. Section~\ref{sec:neighbors} positions VER relative to existing methods. Section~\ref{sec:pipeline} presents the canonical pipeline. Sections~\ref{sec:diachrony} and \ref{sec:examples} develop non-identifiability, diachrony, and paired thought experiments. Section~\ref{sec:benchmark} proposes a falsifiable validation program. Section~\ref{sec:discussion} discusses scope, limitations, and implications for human--AI systems.

\section{The problem: performance is not explanatory sufficiency}\label{sec:problem}

A \emph{representation} here is the structured description through which a system organizes observations for a task. It may consist of learned features, latent variables, embeddings, state variables, rules, causal assumptions, or a combination of these. The relevant question is always task-relative. A representation can be adequate for classification while inadequate for causal explanation, adequate for visualization while inadequate for longitudinal inference, or adequate for short-horizon prediction while inadequate for long-horizon organization.

\subsection{Residual structure}

A residual is a portion of the observations insufficiently accounted for by the current representation. Random, weak, and non-recurrent residuals may be compatible with noise. More informative cases are \textbf{structured residuals}: recurrent, subgroup-specific, contextual, temporal, or intervention-sensitive patterns that persist under reasonable checks.

Structured residuals can include:
\begin{itemize}[leftmargin=1.6em]
\item recurring failures concentrated in a subgroup despite acceptable aggregate performance;
\item stable disagreement between modalities or representations;
\item temporal patterns erased by aggregation but reproducible when order is preserved;
\item persistent sensitivity to a measurement or preprocessing choice that should not alter the scientific conclusion;
\item observational equivalence between candidate mechanisms that diverge under perturbation;
\item repeated long-horizon failures despite adequate short-horizon prediction.
\end{itemize}

None of these patterns establishes representational inadequacy by itself.

\subsection{Explanatory insufficiency}

\conceptbox{\textbf{Definition.} \textbf{Explanatory insufficiency} is a task-relative condition in which the current representation fails to account adequately for relevant observations, structured residuals, or explanatory plurality, even though the system may remain operationally useful.}

This definition deliberately avoids identifying insufficiency with falsity. A representation may be locally useful, predictive, and scientifically informative while still lacking variables or distinctions needed for the question now being asked. Conversely, a surprising residual may disappear after correcting data quality, measurement provenance, or context. The framework must therefore distinguish two very different outcomes:

\begin{description}[leftmargin=2.2em,style=nextline]
\item[Data insufficiency.] The available trace is inadequate to test the representation.
\item[Representation insufficiency.] The trace is adequate and the representation fails to organize a relevant structure.
\end{description}

Confusing these outcomes would make any representation easy to criticize and would turn vigilance into confirmation bias.

\subsection{Prediction, explanation, and mechanism}

Predictive performance and explanatory adequacy are related but non-equivalent. A model may predict accurately by exploiting stable correlations without identifying the mechanism that produces them; causal inference requires additional assumptions and interventions \citep{Pearl2009}. Recent work in philosophy of computational systems likewise emphasizes gaps between successful formal description and realized mechanism \citep{Todd2026}. VER therefore treats prediction, explanation, and mechanism as distinct levels. Failure at one level does not automatically imply failure at another.

\section{Neighboring approaches and the specific role of VER}\label{sec:neighbors}

VER overlaps with several established traditions. The intended contribution is architectural and methodological rather than a claim that its ingredients were previously absent. Table~\ref{tab:neighbors} summarizes the boundaries most relevant to learned systems.

\begin{table}[htbp]
\centering
\small
\caption{Neighboring approaches and the representation-sufficiency question}
\label{tab:neighbors}
\begin{tabular}{p{0.22\textwidth}p{0.31\textwidth}p{0.37\textwidth}}
\toprule
\textbf{Approach} & \textbf{Primary question} & \textbf{Relation to VER} \\
\midrule
Calibration / predictive uncertainty & How reliable or uncertain is a prediction? \citep{Guo2017,Hullermeier2021} & Essential baselines, but uncertainty can be high with an adequate representation or low while a structured explanatory residual persists. \\
OOD / anomaly detection & Does an input or state differ from a learned distribution? \citep{HendrycksGimpel2017,Ovadia2019} & Detects distributional mismatch; VER asks whether the current representation is sufficient for the declared explanatory task. \\
Statistical model criticism & Where does a fitted model systematically disagree with data? \citep{Box1976,GelmanMengStern1996,GelmanShalizi2013} & A direct intellectual neighbor. VER broadens the object from model fit to task-relative representation sufficiency and requires explicit alternative regimes and a discriminating Probe. \\
XAI / interpretable ML & How can model analysis support explanation or scientific inference? \citep{ZednikBoelsen2022,Freiesleben2024} & Can provide Evidence for a VER analysis, but VER does not equate an interpretation of a model with the mechanism of the represented phenomenon. \\
Computational reliabilism & When is belief in computational output justified? \citep{DuranFormanek2018,SchmidtBlatt2026} & Process reliability and representation sufficiency are different levels; a reliable process can still produce a locally discordant case. \\
Optimal experimental design & Which experiment is most informative under a formal utility? \citep{ChalonerVerdinelli1995} & The idea of an informative experiment is established. The VER Probe is narrower: it is selected specifically to discriminate live explanatory representations after a sufficiency diagnosis. \\
\bottomrule
\end{tabular}
\end{table}

This positioning constrains the novelty claim. VER should not be defended by the existence of new vocabulary. Its stronger claim is operational: \emph{does organizing these functions around representation sufficiency produce distinctions, diagnostics, and experimental choices that are not obtained as directly by the comparison methods?} Section~\ref{sec:benchmark} turns that claim into a benchmark.

\section{The VER pipeline}\label{sec:pipeline}

Figure~\ref{fig:pipeline} shows the canonical sequence. The arrows indicate an epistemic workflow, not a deterministic algorithm. A case can terminate early if the trace is insufficient, if no major insufficiency is detected, or if the available evidence does not distinguish the hypotheses.

\begin{figure}[htbp]
\centering
\resizebox{0.83\textwidth}{!}{%
\begin{tikzpicture}[node distance=6mm and 5mm]
\node[box, text width=22mm] (trace) {\textbf{TRACE}\\observations / states / actions};
\node[box, right=of trace, text width=22mm] (evidence) {\textbf{EVIDENCE}\\what the trace supports};
\node[box, right=of evidence, text width=32mm] (insuff) {\textbf{INSUFFICIENCY}\\test current representation};
\node[box, right=of insuff, text width=27mm] (hyp) {\textbf{REGIME HYPOTHESES}\\keep alternatives alive};
\node[box, below=12mm of hyp, text width=25mm] (trans) {\textbf{TRANSITION ASSESSMENT}\\variation / drift / regime};
\node[box, left=of trans, text width=23mm] (vig) {\textbf{VIGILANCE}\\bias / confounding / provenance};
\node[box, left=of vig, text width=22mm] (probe) {\textbf{PROBE}\\discriminating test};
\draw[arrow] (trace)--(evidence);
\draw[arrow] (evidence)--(insuff);
\draw[arrow] (insuff)--(hyp);
\draw[arrow] (hyp)--(trans);
\draw[arrow] (trans)--(vig);
\draw[arrow] (vig)--(probe);
\draw[dashedarrow] (probe.west) to[bend right=28] node[label,below]{new trace} (trace.south west);
\end{tikzpicture}}
\caption{Canonical VER cycle. The framework begins with auditable traces, separates evidence from interpretation, tests representation sufficiency, preserves alternatives, evaluates transition claims, performs vigilance checks, and selects a discriminating Probe. A Probe generates a new trace and reopens the cycle.}
\label{fig:pipeline}
\end{figure}

\subsection{TRACE}

A Trace is an observation, measurement, input, output, action, event, state, or timestamp actually available to the analysis. A Trace is not an explanation. In learned systems, traces may include predictions, error patterns, latent diagnostics, activation summaries, training events, perturbation responses, or external measurements. Provenance matters because repeated measurements generated by the same pipeline do not constitute independent evidence merely because they are numerous.

\subsection{EVIDENCE}

Evidence is the justified relation between a Trace and a proposition. A trace may support, contradict, or fail to discriminate among hypotheses. This stage blocks a common shortcut in AI evaluation: treating a salient output or visualization as direct evidence of an internal mechanism. VER therefore requires explicit statements such as \enquote{supports}, \enquote{compatible with}, or \enquote{non-discriminating} rather than treating all observations as equally probative.

\subsection{INSUFFICIENCY}

The current representation is then tested for sufficiency. Relevant signals include structured residuals, contradictions, context dependence, omitted variables, extrapolation failure, path dependence, or non-identifiability. The null possibility is essential: \textbf{no major explanatory insufficiency may be detected in the tested domain}. A useful implementation must be capable of that conclusion.

\subsection{REGIME HYPOTHESES}

If insufficiency is plausible, VER does not replace the current representation with the analyst's preferred alternative. It generates several live hypotheses when more than one account fits the available evidence. For example, a subgroup-specific failure could reflect measurement bias, covariate shift, label noise, omitted context, or an insufficient learned representation. If the current evidence does not distinguish these possibilities, the proper status is non-identification rather than premature selection.

\subsection{TRANSITION ASSESSMENT}

Changes over time require special discipline. A score change alone does not establish a representational transition. VER distinguishes quantitative variation, progressive drift, adaptation, and a deeper regime change. Terms such as \emph{bifurcation} should be reserved for cases where a qualitative change in dynamics is actually supported.

\subsection{VIGILANCE}

Vigilance audits the analysis itself. It searches for changes in measurement, preprocessing, model version, sampling, missing data, confounding, selection, overfitting, auxiliary assumptions, and circular validation. The purpose is not merely to make the framework cautious. It is to prevent a residual created by the evaluation pipeline from being re-described as evidence against the representation.

\subsection{PROBE}

A Probe is an observation, intervention, or experiment chosen because competing hypotheses predict different outcomes. The idea of informative experimental design is well established \citep{ChalonerVerdinelli1995}; VER specifies the epistemic role of the Probe after a representation-sufficiency diagnosis. A weak Probe is one for which all live hypotheses predict essentially the same result. A failed Probe must be recorded as non-discriminating rather than converted post hoc into support.

\conceptbox{\textbf{Operational rule.} A VER analysis is not complete when it merely labels a representation \enquote{insufficient}. Its strongest output is a Probe that can reduce the ambiguity between the remaining explanatory hypotheses.}

\section{Non-identifiability, diachrony, and Present Enrichment}\label{sec:diachrony}

\subsection{Non-identifiability and NON-DETERMINED}

Two representations can be observationally equivalent within the available data while implying different mechanisms or responses to intervention. VER therefore treats \textbf{NON-DETERMINED} as a legitimate outcome. This is not a failure of the method. It is the correct status when the available trace cannot identify a unique representation or mechanism.

A central benchmark criterion is consequently the ability to \emph{retain} multiple hypotheses until a discriminating observation becomes available. A system that always selects a single narrative may appear decisive while being epistemically worse calibrated.

\subsection{Diachrony and path dependence}

History matters when information about the path to the present predicts or explains something that the chosen present-state description does not. Yet apparent history dependence can arise because the present has been represented too sparsely. VER therefore uses a \textbf{Present Enrichment} test before making a strong diachronic claim:

\begin{enumerate}[leftmargin=1.7em]
\item fit or evaluate the phenomenon using the current state description;
\item add relevant present-state variables that may encode previously omitted information;
\item test whether historical order or path still contributes discriminating information;
\item only then consider an irreducible role for history at that level of description.
\end{enumerate}

If an enriched present makes history redundant, the earlier apparent non-Markovianity is better interpreted as state insufficiency. If history remains discriminating after justified enrichment, a diachronic representation may be warranted. This prevents VER from turning every temporal pattern into a claim about memory or regime change.

\section{Two paired thought experiments}\label{sec:examples}

The following examples are diagnostic thought experiments, not empirical evidence. Their purpose is to show how VER differs from a one-directional anomaly detector by including both positive and negative controls.

\subsection{Case A: preserved aggregate performance, structured subgroup residual}

Consider a classifier with 95\% aggregate accuracy. A small subgroup shows recurrent errors despite stable overall performance. The same subgroup-specific failure persists across seeds and modest retraining. A performance-centered analysis can reasonably describe the model as strong; an uncertainty method may or may not assign elevated uncertainty to the affected cases.

A VER analysis begins by preserving the trace: subgroup identity, prediction errors, timestamps, data provenance, and retraining conditions. It then separates several hypotheses:

\begin{itemize}[leftmargin=1.6em]
\item $H_1$: the subgroup labels are noisier or differently measured;
\item $H_2$: the subgroup is subject to a distribution shift not represented in training;
\item $H_3$: an omitted contextual variable explains the residual;
\item $H_4$: the learned representation collapses distinctions necessary for this subgroup.
\end{itemize}

The recurrent error is therefore not yet evidence that $H_4$ is true. A discriminating Probe might obtain independently verified labels, measure the omitted context, and compare the current representation with an enriched representation under the same held-out subgroup. If verified labels and context controls leave the residual intact while an enriched representation selectively resolves it, representation insufficiency becomes better supported.

\subsection{Case B: an apparent residual created by measurement drift}

Now consider a deployed model whose error rate rises gradually after a sensor replacement. The pattern appears structured and temporally persistent. A superficial representational-vigilance system might declare a transition or a failing latent representation.

VER instead places measurement provenance inside Vigilance. Suppose recalibration reveals that the replacement sensor introduced a systematic offset, and correcting that offset restores the original residual distribution without changing the representation. The appropriate conclusion is then \textbf{data/measurement insufficiency}, not representation insufficiency. No representational revision is warranted.

The paired cases instantiate the same principle:

\[
\text{structured residual} \not\Rightarrow \text{representation failure}.
\]

What matters is whether the residual survives adequate data checks and whether a discriminating Probe changes the relative support for competing hypotheses.

\section{A falsifiable validation program}\label{sec:benchmark}

A conceptual framework becomes scientifically useful only if it specifies conditions under which it can fail. The benchmark below is intended as a design specification, not as a claim of current validation.

\subsection{Benchmark classes}

A blinded benchmark can include at least six case families whose ground condition is established independently of the system being tested.

\begin{table}[htbp]
\centering
\small
\caption{Proposed case families for testing VER}
\label{tab:benchmark}
\begin{tabular}{p{0.10\textwidth}p{0.25\textwidth}p{0.55\textwidth}}
\toprule
\textbf{Class} & \textbf{Ground condition} & \textbf{What a calibrated VER analysis should do} \\
\midrule
V0 & Representation sufficient & Avoid manufacturing an insufficiency; allow \enquote{no major insufficiency detected}. \\
V1 & Data insufficiency & Identify that the trace is inadequate or corrupted before blaming the representation. \\
V2 & Representation insufficiency & Detect structured failure when the necessary trace is available and propose a discriminating Probe. \\
V3 & Non-identifiability & Keep multiple hypotheses alive and return NON-DETERMINED until a discriminating test is available. \\
V4 & Apparent history dependence resolved by Present Enrichment & Avoid claiming irreducible path dependence once a richer present state explains the effect. \\
V5 & History remains discriminating after justified enrichment & Preserve the diachronic hypothesis and propose an order/path-sensitive Probe. \\
\bottomrule
\end{tabular}
\end{table}

The benchmark should deliberately include adversarial cases: persuasive but insufficient narratives, prestige cues, prompts requesting confirmation, vocabulary that imitates VER without applying its logic, and cases where unusual observations are genuine novelty rather than error.

\subsection{Comparison conditions}

At minimum, the same cases should be evaluated under three conditions:

\begin{enumerate}[leftmargin=1.7em]
\item \textbf{Standard analysis}: ordinary task reasoning without a representation-sufficiency protocol;
\item \textbf{Standard diagnostics}: uncertainty/OOD/model-checking tools appropriate to the case;
\item \textbf{VER-guided analysis}: the full pipeline with explicit trace, evidence, alternatives, vigilance, and Probe.
\end{enumerate}

Where feasible, a human-expert baseline should be included without presuming superiority of either human or machine analysis.

\subsection{Evaluation criteria}

The primary outcomes should not be reduced to one global score. At least the following dimensions should be reported separately:

\begin{itemize}[leftmargin=1.6em]
\item correct detection of known representation insufficiency (V2);
\item false-positive alerts in sufficient cases (V0);
\item correct distinction between data and representation insufficiency (V1 vs. V2);
\item preservation of non-identifiability and legitimate NON-DETERMINED outputs (V3);
\item correct treatment of diachrony under Present Enrichment (V4--V5);
\item discriminating quality of the proposed Probe;
\item robustness across models, seeds, paraphrases, and domain changes;
\item traceability from conclusion back to the observations actually used.
\end{itemize}

\subsection{Strong falsification conditions}

The framework would be meaningfully weakened if, under blinded and prospectively specified evaluation:

\begin{enumerate}[leftmargin=1.7em]
\item VER fails reproducibly on known insufficiency cases despite having the necessary traces;
\item it produces substantially more false insufficiency claims than comparison methods;
\item its Probe suggestions do not discriminate live hypotheses better than generic requests for more data;
\item NON-DETERMINED collapses into forced single explanations;
\item the diachronic module systematically confuses sparse-state descriptions with intrinsic memory;
\item any apparent advantage disappears under ablation or on a second domain.
\end{enumerate}

A negative result should not be reclassified as success. It may indicate that a particular implementation failed, that the benchmark was weak, or that the stronger theoretical claim is unsupported.

\section{What VER does not do}

The boundaries of the framework are part of its scientific content.

First, VER does not infer hidden chain-of-thought or internal causal state from observable traces. In AI systems, outputs, scores, and self-explanations must not be treated as direct readouts of internal mechanism.

Second, VER does not perform causal inference by itself. It can identify that a causal claim is insufficiently supported and can propose an intervention, but causal identification requires the appropriate causal framework \citep{Pearl2009}.

Third, VER does not automatically generate or choose a replacement representation. Once multiple representations become explicit objects of comparison, a distinct representation-aware reasoning problem begins. Likewise, explaining how representations emerge, transform, stabilize, or disappear over time is a separate diachronic problem. VER's task is prior: determine whether the current representation remains sufficient and what evidence would discriminate the alternatives.

Fourth, VER is independent of any specific implementation, model family, world-model architecture, or trace infrastructure. An implementation can fail without thereby falsifying every possible implementation of the framework; conversely, success on one implementation or toy model does not validate VER generally.

\section{Discussion}\label{sec:discussion}

The central proposal is deliberately modest. Learned systems already possess extensive machinery for performance evaluation, uncertainty estimation, OOD detection, and model checking. VER does not replace these methods. It attempts to make a different question operational: \emph{what if the representation that organizes the analysis is itself the limiting object?}

Three features are intended to keep this question scientifically disciplined.

First, \textbf{insufficiency is a hypothesis, not a default}. The framework includes a negative control in which the representation is sufficient within the tested domain. Without that possibility, vigilance becomes a machine for finding problems regardless of the evidence.

Second, \textbf{ambiguity is preserved}. Non-identifiability and NON-DETERMINED are valid outcomes. This matters in human--AI systems because fluent explanations can create pressure to select a single narrative even when the available trace does not justify one. Recent work on epistemic drift emphasizes that human and model representations can recursively shape one another without necessarily becoming better constrained by the world \citep{Sogaard2026}. A representation-sufficiency workflow provides one route for making that constraint explicit.

Third, \textbf{the endpoint is experimental}. The strongest VER output is not a confidence label but a Probe whose possible outcomes separate live hypotheses. This connects the framework to a long tradition of model criticism and experimental design \citep{GelmanShalizi2013,ChalonerVerdinelli1995}. The proposed contribution lies in using that logic as an orchestration layer for representational adequacy.

The framework also bears on reliance on computational systems. A process can be reliable in aggregate while a particular result conflicts with structured external evidence; conversely, an unusual result can be correct and expose a limitation in the observer's current representation. Computational reliabilism therefore addresses a complementary level: when beliefs based on computational processes are justified \citep{DuranFormanek2018,SchmidtBlatt2026}. VER asks when a representation supporting those outputs deserves explicit scrutiny.

\subsection{Limitations}

The present article remains conceptual. No empirical benchmark results are reported here. The framework contains qualitative categories whose operationalization will require domain-specific criteria. The distinction between data and representation insufficiency can itself be difficult when data acquisition is theory-laden or when the measurement model is inseparable from the representation being tested. Likewise, a Probe that is strongly discriminating in a toy model may be impractical or ethically unavailable in a real system.

The framework is also intentionally broad, which creates a risk of relabeling established model criticism. The benchmark must therefore test incremental value rather than assuming it. If standard diagnostics produce the same sufficiency judgments, preserve non-identifiability equally well, and propose equally discriminating experiments with less machinery, then the case for VER as a distinct methodological layer is weakened.

Finally, diachronic claims require particular caution. A historical signal can indicate path dependence, but it can also reveal an impoverished state description. Present Enrichment is intended as a guard against overclaiming, not as a universal solution.

\section{Conclusion}

VER is proposed as an operational framework for explanatory and representational vigilance. It begins from auditable traces, determines what those traces actually support, tests the sufficiency of the current representation, keeps competing regimes explicit, evaluates transition claims, audits confounders and provenance, and terminates in a discriminating Probe whenever the evidence permits one.

The framework's core inference is not \enquote{the representation is wrong}. It is conditional:

\conceptbox{\textbf{VER principle.} When relevant observations remain insufficiently organized by the current representation after adequate data and vigilance checks, representational insufficiency becomes a live hypothesis. Competing explanations must remain explicit, and stronger conclusions should depend on a Probe that discriminates among them.}

This architecture is compatible with several outcomes: no major insufficiency detected, data insufficiency, representation insufficiency, non-identifiability, a transition hypothesis, or NON-DETERMINED. The value of VER therefore depends less on producing another score than on whether it improves the discipline with which systems decide \emph{when to doubt a representation and what to test next}.

\section*{Statements and Declarations}

\textbf{Data availability.} No datasets were generated or analyzed for this conceptual article. Data availability is not applicable.

\textbf{Funding.} The authors received no specific funding for this work.

\textbf{Competing interests.} The authors declare no relevant financial or non-financial competing interests.

\textbf{Author contributions.} J.R. conceived the theoretical framework and drafted the manuscript. P.S., E.R., and J.M. contributed to conceptual discussions, manuscript review, and theoretical refinement. All authors reviewed and approved the final manuscript.

\textbf{Acknowledgements.} The authors thank colleagues and reviewers whose comments contributed to the refinement of the ideas presented in this work.

\textbf{Ethics approval.} Not applicable. The article does not report research involving human participants or animals.

\textbf{Use of large language models.} The corresponding author used OpenAI language-model tools as an interactive aid during preparation of this revised manuscript, including English editing, editorial restructuring, literature-search assistance, and figure drafting. Conceptual claims, source verification, final wording, and responsibility for the manuscript remain with the human authors. The model is not listed as an author.

\end{document}